\pdfoutput=1
\documentclass{article}

\PassOptionsToPackage{numbers, compress}{natbib}
\usepackage[preprint]{neurips_2024}

\usepackage[utf8]{inputenc} % allow utf-8 input
\usepackage[T1]{fontenc}    % use 8-bit T1 fonts
\usepackage{hyperref}       % hyperlinks
\usepackage{url}            % simple URL typesetting
\usepackage{booktabs}       % professional-quality tables
\usepackage{amsfonts}       % blackboard math symbols
\usepackage{nicefrac}       % compact symbols for 1/2, etc.
\usepackage{microtype}      % microtypography
\usepackage{xcolor}         % colors
\usepackage{xspace}
\usepackage{amsmath}
\usepackage{pifont}% http://ctan.org/pkg/pifont
\usepackage{graphicx}
\usepackage{enumitem}
\usepackage{subfig}
\usepackage{multirow}
\usepackage{wrapfig,lipsum,booktabs}
\usepackage[capitalise]{cleveref}
\newcommand{\xmark}{\ding{55}}%
\usepackage{enumitem}
\newcommand{\OURS}{FaceGPT\xspace}

\title{\OURS: Self-supervised Learning to Chat\\ about 3D Human Faces}

\author{%
Haoran Wang\textsuperscript{$1$}
\enskip 
 Mohit Mendiratta\textsuperscript{$1$}  \enskip 
Christian Theobalt \textsuperscript{$1$}
\enskip \textbf{Adam Kortylewski}\textsuperscript{$1,2$} \\ 
{\textsuperscript{$1$} Max Planck Institute for Informatics    } \enspace 
{\textsuperscript{$2$} University of Freiburg} \enspace  \\
}

\begin{document}

\maketitle

\begin{abstract}
We introduce \OURS, a self-supervised learning framework for Large Vision-Language Models (VLMs) to reason about 3D human faces from images and text. Typical 3D face reconstruction methods are specialized algorithms that lack semantic reasoning capabilities. \OURS overcomes this limitation by embedding the parameters of a 3D morphable face model (3DMM) into the token space of a VLM, enabling the generation of 3D faces from both textual and visual inputs. \OURS is trained in a self-supervised manner as a model-based autoencoder from in-the-wild images. In particular, the hidden states of LLM is projected into 3DMM parameters and subsequently rendered as 2D face image to guide the self-supervised learning process via image-based reconstruction. Without relying on expensive 3D annotations of human faces, \OURS obtains a detailed understanding about 3D human faces, while preserving the capacity to understand general user instructions. Our experiments demonstrate that \OURS not only achieves high-quality 3D face reconstructions but also retains the ability for general-purpose visual instruction following. Furthermore, \OURS learns fully self-supervised to generate 3D faces based on complex textual inputs, which opens a new direction in human face analysis. 
\end{abstract}

% \section{Submission of papers to NeurIPS 2024}
\section{Introduction}
In this work, we address the problem of reasoning about 3D human faces from images and text. 
Traditional monocular 3D face reconstruction methods typically estimate the parameters of a 3D morphable model \cite{blanz1999morphable,tewari17MoFA,deng2019deep3d,Feng2021deca,li2023focus} given 2D face images as input.
However, these methods lack the capability to reason about faces based on textual input.
Unlike these systems, humans can vividly imagine and even draw faces based solely on either face images or textual descriptions.
Motivated by recent advances in large vision-Language models (VLMs) \cite{liu2023llava, zhu2023minigpt},
we aim to explore a path forward towards enabling vision-language systems to obtain an in-depth reasoning-based understanding of 3D faces.

\begin{figure}
  \centering
  \includegraphics[width=\textwidth]{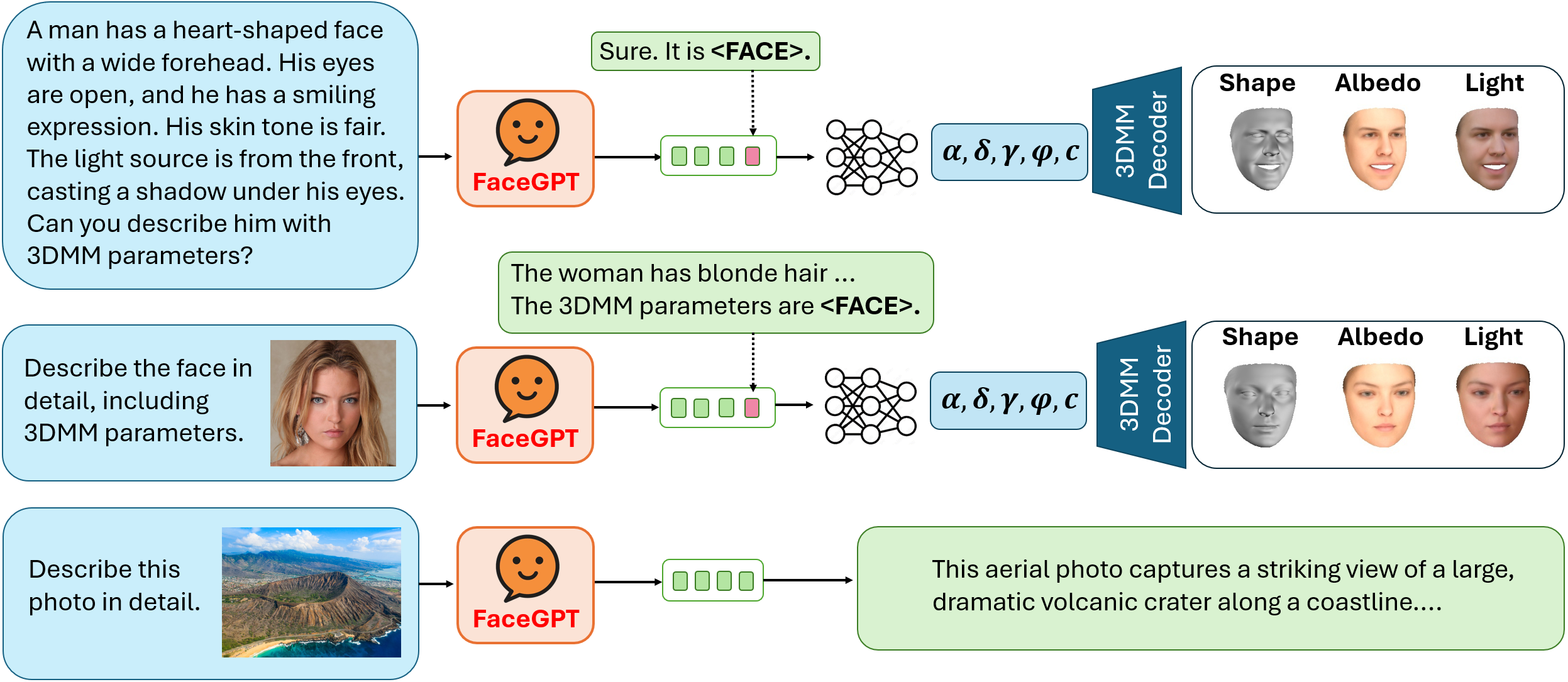}
  \caption{We introduce \OURS, a large vision-language model that learns to produce 3D human faces (in terms of 3DMM parameters) in a fully-self-supervised manner.
  When prompted with face images and task-specific questions, \OURS can output a special \texttt{<FACE>} token of which the corresponding feature embedding (red) can be decoded into 3DMM parameters, that encode the face shape $\alpha$, expression $\delta$, the texture $\gamma$,  the light $\phi$ and camera $c$ parameters.
  When decoded with a 3DMM and differentiable renderer, this enables a fully self-supervised learning via inverse rendering. 
  \OURS can produce 3D human faces from text-only input (first row), as well as multi-modal input (second row) without requiring any 3D annotations, while retaining general chatting abilities (last row).}
  \label{fig:teaser}
\end{figure}

To investigate this question, we present \OURS, a vision-language model with an intricate ability to reason about 3D human faces from visual and textual input.
We represent faces as 3D morphable model (3DMM) parameters that include parameters for the 3D face shape, expression, albedo, and scene illumination.
Following related work on image segmentation \cite{lai2023lisa} and human pose estimation \cite{feng2024chatpose}, we extend the original token space of the VLM by incorporating a new \texttt{<FACE>} token that is decoded into 3DMM parameters using an MLP (\cref{fig:teaser}).
Thus, when combined with a differentiable computer graphics renderer \cite{ravi2020accelerating}, the VLM model becomes capable to synthesize face images.
This enables us to formulate \OURS within a model-based autoencoder framework \cite{tewari17MoFA}, and hence to train our model in a fully self-supervised manner from in-the-wild images.
To the best of our knowledge, this is the first work combining vision-language model with an inverse graphics pipeline.
During training, we freeze the visual encoder of the VLM while training the MLP and the LLM using LORA \cite{hu2022lora}. 
The model is trained with three types of data: (1) Standard multi-modal instruction tuning data to retain the general capability and quality of the VLM. (2) In-the-wild face images for training the \texttt{<face>} token and 3DMM projection layers via self-supervised inverse rendering. (3) Text-to-3DMM data that enables the generation of 3DMM parameters from text input only. We construct this dataset from a set of face images in the wild by running an off-the-shelf self-supervised monocular face reconstruction method \cite{li2023focus} and by generating textual descriptions of the faces via the original VLM.

We evaluate \OURS on a variety of tasks, including traditional 3D face reconstruction, text-to-3DMM face generation and general-purpose visual instruction following.
We demonstrate that \OURS becomes a general-purpose model that achieves competitive results at 3D face reconstruction when compared to specialized methods, while retaining competitive visual instruction following capabilities, as well as being capable of reasoning-based text-to-3D face generation. 
Beyond face analysis, we believe that the design principles underlying \OURS are general and it also suggests a pathway towards a self-supervised integration of the "world knowledge" that VLMs derive from extensive textual data and the structured 3D representations of the visual world via computer graphics.

In summary, our work makes the following concrete contributions:
\begin{itemize}
\item To the best of our knowledge, this is the first work that enables vision-language models to learn a detailed 3D face understanding in a fully self-supervised manner.
\item We show that VLMs can learn text-based face reconstruction, which predicts 3D human faces given user instructions, in a fully self-supervised manner. 
\item Our experiments on traditional 3D face reconstruction, visual instruction following and text-based face reconstruction, demonstrate the general face understanding capabilities of \OURS.
\end{itemize}

\section{Related works}
\subsection{Monocular model-based face reconstruction}
Realistically reconstructing digital human faces has been a longstanding challenge in computer vision and graphics due to their vast potential applications. Traditional methods primarily use parametric 3D Morphable Models (3DMM)~\cite{blanz1999morphable,paysan20093d,li2017learning} with PCA for dimensionality reduction to simplify high-dimensional 3D face scans, serving as a 3D prior for representing unique facial characteristics and providing precise control. Recently,  deep learning-based methods that map 2D images to 3D face models have gained popularity. Early methods struggled with the need for extensive 3D facial scan data paired with 2D images, which was labor-intensive and costly. This limitation was addressed with the introduction of model-based face autoencoders (MoFA) \cite{tewari17MoFA} that enabled self-supervised 3D face reconstruction. 
MoFA uses a differentiable rendering layer to minimize differences between input and rendered images, enabling end-to-end learning without ground-truth 3D faces, leading to a number of effective extensions in the self-supervised learning strategies ~\cite{tewari2018self,tewari2018high,bas20173d,genova2018unsupervised,danvevcek2022emoca}.
RingNet~\cite{sanyal2019learning} and DECA~\cite{feng2021learning} use landmark-based training, predicting landmarks for input images and treating them as pseudo ground truth, measuring the distance between 2D face landmarks and their projections on the 3DMM surface.
The FOCUS~\cite{li2023focus} framework jointly trains a face autoencoder and an outlier segmentation network, which makes the method robust to outliers such as occlusion and make-up. These advancements significantly improved monocular model-based face reconstruction, making it more efficient and effective.
However, these methods are highly specialized and lack a deep understanding of the semantics of human faces or the ability to relate faces to language, limiting their overall scope and effectiveness.
\subsection{Text-to-3D Face Generation and Manipulation}
Text-to-3D face generation and manipulation methods aim to integrate textual information for creating and editing 3D faces. 
Methods like Dreamface~\cite{zhang2023dreamface} and Describe3D~\cite{wu2023high} generate text-conditioned texture maps to render 3D morphable models (3DMM).
TG-3DFace~\cite{yu2023towards} advances this by using tri-plane neural representations and extending the 3D-aware GAN, EG3D~\cite{chan2022efficient}, for end-to-end text-conditioned generation. It incorporates a face parser and attribute classifier for fine-grained semantics and uses a camera pose conditional discriminator to guide learning. For text-guided 3D face manipulation, methods like Latent3D~\cite{canfes2022text} and ClipFace~\cite{aneja2023clipface} optimize intermediate layers with a CLIP-based loss to generate UV-texture maps or predict texture and expression latent codes. 
These methods, however, rely on 3D scan data and to train new mappers for each text instruction.
Unlike these task-specific approaches, \OURS is a general-purpose model that reasons about 3D human faces from images, text, or both by leveraging general visual knowledge. Our model can interact with users through conversations, discussing facial features and providing relevant responses, while also being capable of following general user instructions.

\subsection{Multimodal Large Language Models}

Large Language Models (LLMs) are rapidly transforming various fields~\cite{radford2019gpt2, NEURIPS2020gpt3, chatgpt, gpt4}. While proprietary models like OpenAI’s ChatGPT~\cite{chatgpt} and GPT-4~\cite{gpt4} dominate the landscape, open-source alternatives such as Vicuna~\cite{chiang2023vicuna}, LLaMA~\cite{touvron2023llama}, and Alpaca~\cite{taori2023alpaca} support research efforts. However, LLMs mainly focus on generating text as output given text-only input. The integration of additional modalities into LLMs represents an active area of research.

Multi-Modal Large Language Models (MM-LLMs) are emerging, extending LLMs' capabilities beyond text to encompass a broader spectrum of modalities, including images, videos, and audio. In the realm of image-text understanding, recent endeavors like LLaVA~\cite{liu2024visual} and MiniGPT-4~\cite{zhu2023minigpt} incorporate vision encoders to interpret images and align their features with language embeddings using projection layers. Moreover, cutting-edge models such as PandaGPT~\cite{su2023pandagpt}, ImageBind~\cite{girdhar2023imagebind}, and NeXT-GPT~\cite{wu2023next} exhibit impressive versatility in handling diverse modalities, aligning embeddings from text, images, audio, and video with language as both input and output. To enhance LLM to natively output more modalities, approaches like LISA~\cite{lai2023lisa} connects LLaVA with a decoder to generate text and segmentation masks, while ChatPose~\cite{feng2024chatpose} specializes in human pose information. However, these methods typically rely on supervised learning.

In our study, we explore the integration of human faces as a novel modality for LLMs. We assess (1) LLMs' ability to generate 3D faces from text or image inputs and (2) their capability to understand human facial features and integrate this understanding into their broader functionality without supervision.

\section{Method}
\begin{figure}
  \centering
  \includegraphics[width=\textwidth]{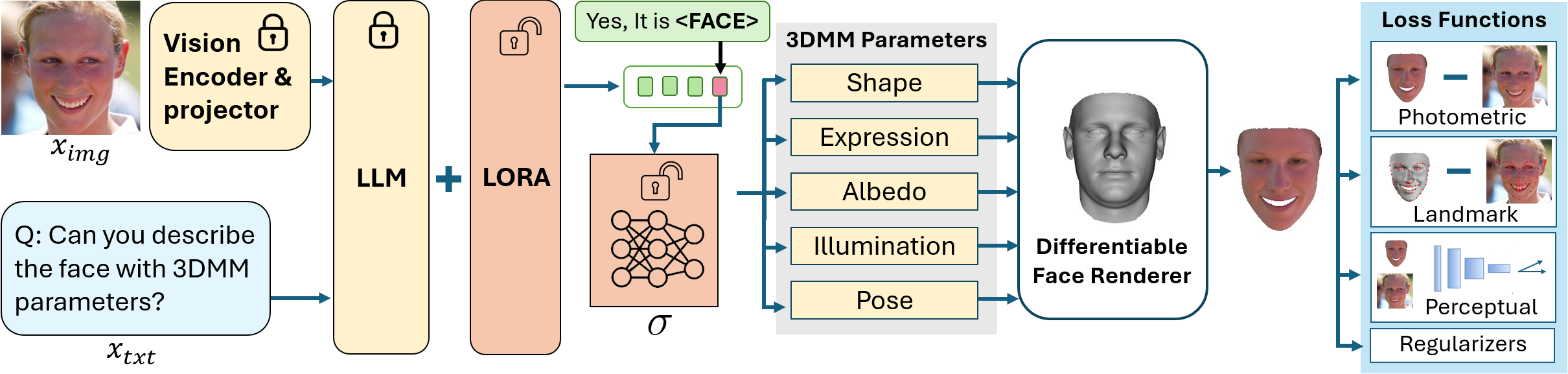}
  \caption{Architecture of FaceGPT. Our model consists of a multi-modal LLM, which includes a vision encoder, a vision projection layer, and the LLM itself, along with a 3DMM projection layer, denoted as $\sigma$, and the parametric Basel face model~\cite{blanz1999morphable}. During the training phase, we focus on training the $\sigma$ projection layer and fine-tuning the LLM, while keeping other components frozen. The network components are trained using a differentiable renderer to compute a self-supervised loss.} 
  \label{fig:architecture}
\end{figure}

In this work, our aim is to augment existing Large Vision Language Models (VLMs) with the ability of reasoning about 3D human faces without requiring manual human efforts. 
Inspired by established model-based face reconstruction methods, we treat the 3D human face as a separate modality within the LLM framework and represent it with 3D Morphable Model (3DMM) \cite{blanz1999morphable}parameters that represent the face shape, expression, albedo and illumination. 
In particular, we  introduce a \texttt{<FACE>} token into the token space of the LLM, which is mapped into the 3DMM parameter space and subsequently rendered into a 2D image, hence enabling self-supervised 3D facial reconstruction.

\subsection{Model Architecture}
\cref{fig:architecture} illustrates 
the \OURS architecture.
The VLM has an extended vocabulary which includes a new token $\texttt{<FACE>}$ that specifically represents the human face. Given an input image $x_{img}$ and/or text prompt $x_{txt}$, the VLM $f$ predicts text responses:
\begin{equation}
{y}_{txt} = f(x_{img}, x_{txt}),
\end{equation}
where ${y}_{txt} =[t_1,\dots,t_N]$ is the output sequence of tokens with corresponding hidden states $[h_1,\dots,h_N]$.
If one of the output tokens $t_n\in y_{txt}$ is our newly defined \texttt{<FACE>} token, we can extract the hidden state as $h_{\texttt{<FACE>}}=h_n \in \mathbb{R}^{4096}$ and project it using an MLP $\sigma$ into the latent 3DMM parameters $\theta=\sigma(h_{\texttt{<FACE>}})= [\alpha, \delta, \gamma,  \phi, c] \in \mathbb{R}^{257}$,
 i.e. the 3D face shape $\alpha \in \mathbb{R}^{80}$, facial expression parameters $\delta \in \mathbb{R}^{64}$ and texture $\gamma \in \mathbb{R}^{80} $ of a 3DMM, as well as the spherical harmonics illumination $\phi  \in \mathbb{R}^{27} $ and camera parameters $c \in \mathbb{R}^{6}$ of the scene.
The 3D vertices and triangles, as well as the color of the face mesh are then determined using the standard 3DMM model $M(\theta)$ as described in \cite{Feng2021deca}. 
Using an orthographic camera model the reconstructed 3D face mesh can be rendered into 2D space using a differentiable renderer $\Pi$, hence producing the final reconstructed face image $\hat{y}_{rec}$. The  process can be summarized as:
\begin{subequations}
  \begin{align}
    \theta & = \sigma(h_{\texttt{<FACE>}})\\
    %V, R & = MoFA(\theta)\\
    \hat{y}_{rec} & = \Pi(M(\theta))
  \end{align}
\end{subequations}
Note that the \OURS architecture can be seen as a new type of language-based autoencoder, with a VLM as encoder and a computer graphics decoder that is based on the 3DMM.

\subsection{Self-supervised Training}
\label{subsec:training_loss}

Our objective is to develop a VLM that is able to learn an enhanced comprehension of 3D human faces in a self-supervised manner. 
Ultimately, the model should not only understand user instructions reliably, but also to accurately reconstruct 3D faces from visual or text input.
To accomplish this, we have devised a self-supervised approach that incorporates 3DMM understanding into existing VLMs. 
This method allows us to leverage face data without the need for costly 3DMM annotations. 
We also construct a text-to-3DMM dataset in an unsupervised manner that supports this training paradigm, enabling our model to learn these capabilities effectively and efficiently.

\textbf{Training Objective.} Inspired by the success of self-supervised learning in LLMs \cite{radford2019gpt2}, we augment the existing VLM training objective with a self-supervised face reconstruction loss $L_{face}$ that avoids the expensive 3D annotation of human faces in in-the-wild images via image-based reconstruction to regress 3DMM parameters:
\begin{equation}
    L=\lambda_{txt}L_{txt} + L_{face}
\end{equation}
where $\lambda_{txt}$ and $\lambda_{face}$ are the weights for balancing the losses.

\textbf{Text prediction loss.} 
We utilize the autoregressive objective $L_{txt}$ to guide the model to generate correct text output and hence retain its general capability to follow user instructions:
\begin{equation}
        L_{txt} = CE(\hat{y}_{txt}, y_{txt}),
\end{equation}
where $y_{txt}$ is the ground truth text response and $CE$ is the cross entropy loss.

\textbf{Self-supervised face reconstruction loss.} 
To incorporate the 3DMM as a new modality into an existing VLM, we add a new token $\texttt{<FACE>}$ in the vocabulary of the LLM and fine-tune the language modelling head. 
For fine-tuning the VLM without relying on manually annotated data, we introduce a 2D self-supervised loss $L_{face}$, following established protocols of specialized face reconstruction models \cite{li2023focus}:
\begin{equation}\label{eq:face_reconstruction}
L_{face} = \lambda_{pixel}L_{pixel} + \lambda_{per}L_{perc} + \lambda_{LM}L_{LM} 
+ \lambda_{reg}L_{reg}, 
\end{equation}
where $L_{pixel} = ||A\odot(x_{img}-\hat{y}_{rec})||^2_2$ refers to the pixel-wise reconstruction loss between reconstructed images $\hat{y}_{rec}$ and the input images $x_{img}$. 
To avoid distortions when training from in-the-wild data, a 2D skin mask $A$ is estimated from the input images using a simple pre-trained Gaussian mixture model for the skin color \cite{deng2019deep3d}.

The perceptual loss $ L_{perc}   = sim(f_{perc}(x_{img}), f_{perc}(\hat{y}_{rec}))$ estimates the cosine similarity between the reconstructed image $\hat{y}_{rec}$ and the input image $x_{img}$ at the perceptual level using a pre-trained feature extractor $f_{perc}$.

The landmark loss $L_{LM} = ||LM_{img}-LM_{rec}||^2_2$ measures the L2 distance between the projected 2D landmarks of the estimated 3DMM $LM_{rec}$ and predicted 2D facial landmarks $LM_{img}$ using an off-the-shelf facial detector \citep{bulat2017far}.

$L_{reg} = ||\theta||^2_2$ regularizes the estimated 3DMM parameters towards the mean value of the multi-variate Gaussian distribution of the 3DMM. 

The weight parameters $\lambda_{face}= [ \lambda_{pixel}, \lambda_{per}, \lambda_{LM}, \lambda_{reg}]$ balance the respective losses, and we set them as established in prior work \cite{li2023focus}. 

\textit{Preserving the ability for natural conversations about faces.} We noticed that VLM's loose the ability to conduct general conversations about faces when trained self-supervised face reconstruction loss, and tend to always output 3DMM parameters when queried with a face image. 
To resolve this problem, we generate a face conversation dataset with accurate textual face descriptions, by querying the VLM with face images and conversations that include multiple questions about facial attributes.
During training, we mix task-specific instructions that explicitly ask for 3DMM parameters with general conversational data to regularize the training and preserve the ability for general non-3DMM related conversations about faces.
In contrast to prior works \citep{lai2023lisa, feng2024chatpose} which generally employ a simple static question-answer template that restrains the model from providing diverse responses, our proposed strategy can effectively improve the instruction following performance for general face conversations.

\textbf{Self-supervised Text-to-3DMM loss.}
We aim to enable the VLM to perform semantic face understanding tasks, i.e. to predict faithful 3DMM parameters when only having textual input $f(\cdot,x_{txt})$.
While related works for 3D human pose estimation \cite{delmas2022posescript,feng2024chatpose} rely on human-written fine-grained language descriptions of 3D human meshes, we aim to achieve text-to-3DMM generation without any human interaction.
To achieve this, we introduce a self-supervised text-based face reconstruction loss. 
In particular,  we first query the VLM with the face images in the training data and a question template that requests detailed face descriptions of a number of facial attributes such as head shape, skin tone, or facial expressions. 
This results in a dataset with pairs of face images and  detailed textual descriptions about facial attributes.
During training, we instruct the VLM to estimate 3DMM parameters only from the available detailed face descriptions $f(\cdot,x_{txt})$.
As we have the corresponding 2D image for each face description, we can re-use the self-supervised face reconstruction loss \cref{eq:face_reconstruction} to train the model at the task of text-to-3DMM generation, without the need for any human-written annotation.

\section{Experiments}
\label{experiments_part}
% \begin{table}
%   \caption{Sample table title}
%   \label{sample-table}
%   \centering
%   \begin{tabular}{lll}
%     \toprule
%     \multicolumn{2}{c}{Part}                   \\
%     \cmidrule(r){1-2}
%     Name     & Description     & Size ($\mu$m) \\
%     \midrule
%     Dendrite & Input terminal  & $\sim$100     \\
%     Axon     & Output terminal & $\sim$10      \\
%     Soma     & Cell body       & up to $10^6$  \\
%     \bottomrule
%   \end{tabular}
% \end{table}
\subsection{Training Data}
\vspace{-.1cm}
\label{training_data_details}

\textbf{Text-to-Face Data.} The text-to-face dataset comprises pairs of text descriptions and face images, facilitating the development of mappings by VLM between textual descriptions and the 3DMM parameters of faces. During training, only text is used as input and the corresponding face images are only used to produce supervision signals. Given the absence of publicly available datasets linking text descriptions to 3D face meshes and existing VLMs like LLaVA present powerful Visual Question Answering(VQA) ability, we generate textual descriptions for face images. We employ templates structured as follows to guide the learning process in VLMs: \texttt{"USER: \{description\}, can you give the 3DMM parameters of this person. ASSISTANT: Sure, it is <FACE>."}, where \texttt{\{description\}} is the text description for faces. We utilize high-quality face images from CelebA-HQ \citep{karras2017progressive} and generate text descriptions using LLaVA-v1.5-13B \citep{liu2023improvedllava} to construct the dataset. 

\textbf{Image-to-Face Data.} Image-to-face reconstruction data is composed of only human face images. The face images will be formatted with task-specific templates like \texttt{"USER: <IMAGE> Can you give the 3DMM parameters of this person. ASSISTANT: Sure, it is <FACE>."}. To enhance the diversity and relevance of the conversations centered around human face images, we also generate face-centric conversations for each face image. This approach enriches the contextual understanding of the VLM concerning the newly introduced token \texttt{<FACE>}. We adopt the CelebA-HQ trainset as the image to face reconstruction dataset. 

\textbf{Multimodal Instruction-Following Data.} This data is general-purpose VQA data, and it is used to preserve the VLM's ability of understanding a user's instructions. Following LLaVA v1.5, We use LLaVA-v1.5-mix665k as multimodal instruction following data.

\subsection{Experimental settings}
\vspace{-.1cm}
\textbf{Network Architecture.} We build our model on Large Vision Language Model LLaVA-1.5-7B \citep{liu2023improvedllava} with CLIP-ViT-L-336px as vision encoder and Vicuna v1.5 as the LLM backbone. LoRA \citep{hu2022lora} is applied to efficiently fine-tune the VLM. An MLP head with GeLU activations \citep{hendrycksG2016gelu} is appended to the last layer of the VLM to predict the 3D human face parameters.

\textbf{3D Face Model.} We use the Basel Face Model (BFM) 2017 \cite{gerig2018morphable} as the 3D face model. The face is parameterized as the semantic code vector $\theta=[\alpha, \delta, \gamma,  \phi, c]\in \mathbb{R}^{257}$ in \citep{tewari17MoFA}, which includes 3D shape parameters $\alpha\in \mathbb{R}^{80}$, facial expression parameters $\delta\in \mathbb{R}^{64}$, texture of 3DMM $\gamma\in \mathbb{R}^{80}$, illumination $\phi\in \mathbb{R}^{27}$ and camera parameters $c\in \mathbb{R}^{6}$.

\textbf{LLaVa-Key baseline.} As there are no VLM-based methods that can perform 3D face reconstruction, we introduce the LLaVa-Key. The model is finetuned to predict facial landmarks in the format of pure text given either images or textual description of human faces as input. For each input, gradient-based optimization is applied to these predicted landmarks to fit the 3DMM using \cref{eq:face_reconstruction} to obtain a 3D face reconstruction. It is important to note that LLaVa-Key employs test-time fine-tuning, making it an inherently different baseline compared to our approach and other task-specific methods, which rely solely on feed-forward prediction.
Nevertheless, we still use LLaVA-Key as a baseline to illustrate the effect of a direct 3DMM embedding in the token space of \OURS, which generally yields superior results, even without any test-time fine-tuning. 

\textbf{Implementation Details.}
The training uses 8 NVIDIA 48G A40 GPUs. We utilize deepspeed \citep{ras2020deepspeed} engine and ZeRO optimizer \citep{raj2020zeroopt} for efficient training. We use AdamW \citep{loshchilov2018adamw} optimizer with the learning rate and weight decay set to $2e-5$ and 0, respectively. We also follow the standard setting of using a WarmupDecayLR as the learning rate scheduler, where the warm-up iterations are set to 100. The weights of the text generation loss $\lambda_{txt}$ and the face reconstruction loss $\lambda_{face}$ are set to 1.0 and 0.1, respectively. And following \citep{li2023focus},  those of the pixel loss $\lambda_{pixel}$, the perceptual loss $\lambda_{per}$, the landmark loss $\lambda_{LM}$, and the regularization loss $\lambda_{reg}$ are set to 0.5, 0.25, 5e-4 and 0.1, respectively. The landmarks of human faces are obtained with \citep{bulat2017far} and we use pre-trained ArcFace \citep{deng2019arcface} to extract perceptual features. The batch size is set to 8 per GPU and the gradient accumulation step is set to 4.

\textbf{Evaluation Metrics.} 
The instruction following ability is measured with GPT-assisted evaluation, which queries GPT4 to obtain the grading of generated responses. The 3D human face parameters estimation is measured with the L2 photometric error in RGB space and the L2 landmark error.

\begin{figure}
	\centering
	\scalebox{.9}{
		\begin{tabular}{c|ccc|cc}
			\subfloat{\includegraphics[width=0.14\linewidth]{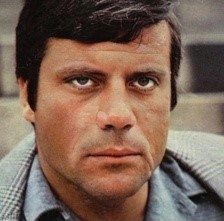}} &
			\subfloat{\includegraphics[width=0.14\linewidth]{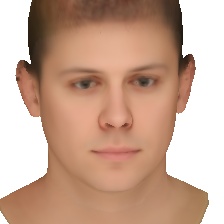}} &
			\subfloat{\includegraphics[width=0.14\linewidth]{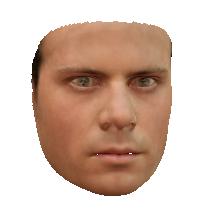}} &
			\subfloat{\includegraphics[width=0.14\linewidth]{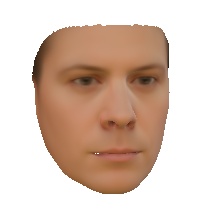}} &
            \subfloat{\includegraphics[width=0.14\linewidth]{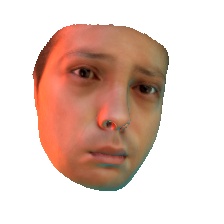}} &
            \subfloat{\includegraphics[width=0.14\linewidth]{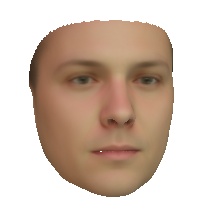}}\\
			
			\subfloat{\includegraphics[width=0.14\linewidth]{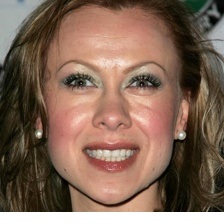}} &
			\subfloat{\includegraphics[width=0.14\linewidth]{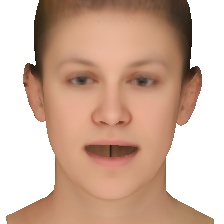}} &
			\subfloat{\includegraphics[width=0.14\linewidth]{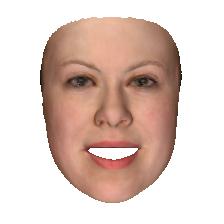}} &
			\subfloat{\includegraphics[width=0.14\linewidth]{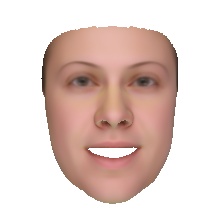}} &
            \subfloat{\includegraphics[width=0.14\linewidth]{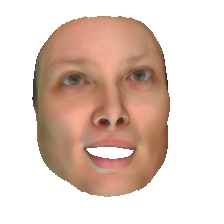}} &
            \subfloat{\includegraphics[width=0.14\linewidth]{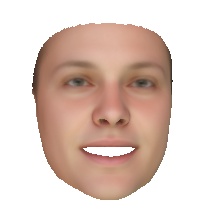}}\\

			\subfloat{\includegraphics[width=0.14\linewidth]{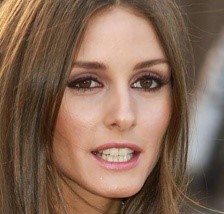}} &
			\subfloat{\includegraphics[width=0.14\linewidth]{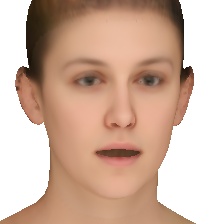}} &
			\subfloat{\includegraphics[width=0.14\linewidth]{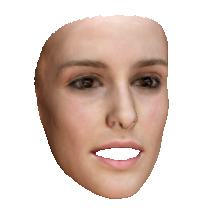}} &
			\subfloat{\includegraphics[width=0.14\linewidth]{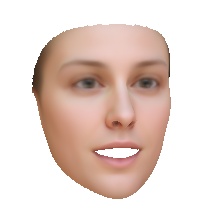}} &
            \subfloat{\includegraphics[width=0.14\linewidth]{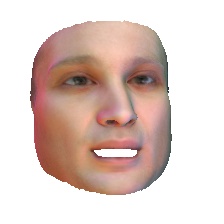}} &
            \subfloat{\includegraphics[width=0.14\linewidth]{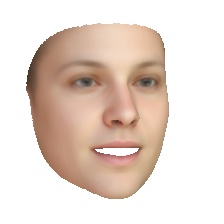}}\\

			Input Image & DECA & Deep3D & FOCUS & LLaVA-Key & \OURS\\
	\end{tabular}
 }
	\caption{Qualitative results for 3D face reconstruction. Our approach allows for the regression of pose, shape, expression, skin reflectance, and illumination from a single monocular image, achieving quality comparable to recent state-of-the-art methods. }
	\label{fig:rec_quali}
\end{figure}

\subsection{Image-based 3D Face Reconstruction}

\begin{wraptable}{r}{5cm}
\caption{Performance at monocular 3D face reconstruction.}
  \label{tab:classic_face_rec_res}
  \centering
  \begin{tabular}{lll}
    \toprule
    Method     & photo     & keypoint      \\
    \midrule
    DECA \citep{Feng2021deca}    &  0.216     &  5.2px      \\
    Deep3D \citep{deng2019deep3d}&  0.073 & 3.2px      \\
    FOCUS \citep{li2023focus}    &   0.077    &  2.2px     \\
    \hline
    LLaVa-Key & 0.110 &  14.0px\\
    FaceGPT    &  0.106     & 3.0px      \\
    \bottomrule
  \end{tabular}
\end{wraptable}

We evaluate \OURS at image-based 3D face reconstruction and show a comparison with SOTA specialized monocular face reconstruction methods and a VLM-based generalist baseline LLaVa-Key in \cref{tab:classic_face_rec_res}. 

Moreover, \cref{fig:rec_quali} shows a qualitative comparison of \OURS and all baselines at the classic task of 3D face reconstruction from a single image.
As reported in prior works for VLM-based segmentation \cite{lai2023lisa} and human pose estimation \cite{feng2024chatpose}, our \OURS model does, as expected, not reach the state-of-the-art performance of specialized models.
However, we note that \OURS has a frozen vision encoder to preserve the generalist behavior, whereas all baseline models have a fine-tuned task-specific backbone.
Moreover, our model does outperform DECA, which is a highly competitive and widely applied baseline model \cite{ding2023diffusionrig}.
When compared to the VLM baseline model, \OURS achieves large improvements, highlighting the benefit of embedding the 3DMM parameters directly in the token space of the VLM.

\subsection{Text-based 3D Face Reconstruction}

\begin{figure}
    \centering    \includegraphics[width=\linewidth]{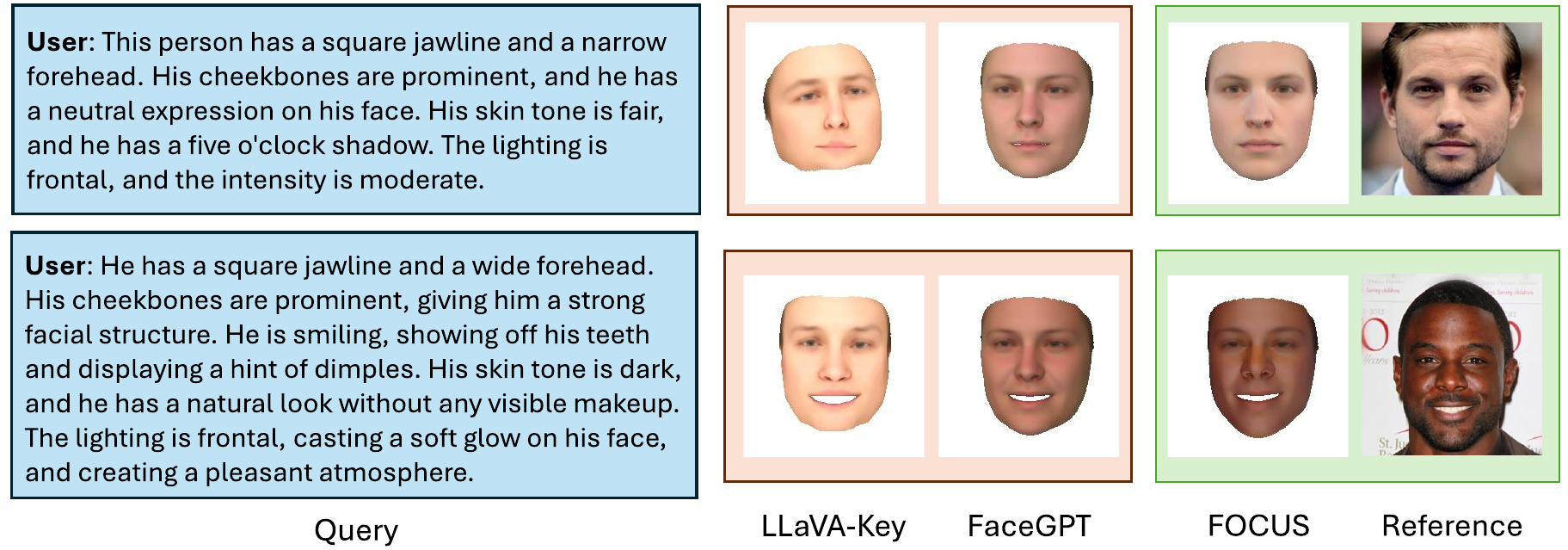}
    \caption{Qualitative results of text-to -3DMM synthesis comparing the performance of the baseline VLM LLaVA-key and FaceGPT. Our method can reason and generate faithful 3D faces using only textual description. We also present the realistic face image which matches the description as a reference. A 3DMM reconstruction of reference images estimated by FOCUS method \citep{li2023focus} is also provided for reference. Our approach captures many facial details from the texts and produces much better visual results compared to the LLaVA-Key baseline method.
    }
    \label{fig:quali-txt23dmm}
\end{figure}

\cref{exp:reason_face_rec_res} and \cref{fig:quali-txt23dmm} show the results of \OURS at text-based 3D face reconstruction. As there are no public unsupervised methods available that perform this task, we compare to a VLM-based baseline LLaVA-Key, where LLaVA is fine-tuned to predict the 2D coordinates of face landmarks given text description as input and an optimization-based method is utilized to fit a 3DMM on predicted landmarks. 
\begin{wraptable}{r}{4.5cm}
\addtolength{\tabcolsep}{-0.4em}
  \caption{Performance evaluation of text-based 3D face reconstruction (as L2 3DMM error).}
  \label{exp:reason_face_rec_res}
  \centering
  \begin{tabular}{llr}
    \toprule
    Method    & Exp.& Shape \\
    \midrule
    LLaVA-Key  &  9.39     &  12.10     \\
    FaceGPT    &  2.89     &  4.04       \\
    \bottomrule
  \end{tabular}
\end{wraptable}
We can clearly observe the significant advantage that \OURS achieves over the baseline method. 
Moreover, the qualitative results demonstrate that the model can achieve faithful text-based 3D face reconstructions, despite being trained in a self-supervised manner. 
These results demonstrate that embedding 3D knowledge about our world into a VLM is in principle possible without detailed human annotations, hence demonstrating the large potential of combining VLM's with vision-as-inverse graphics for self-supervised learning.

\subsection{GPT-Assisted Evaluation}
Table \ref{gpt_res} shows that \OURS preserves the ability of instruction following by following LLaVA’s evaluation \citep{liu2023llava} protocol, using GPT4-Assisted evaluation on LLaVA-Bench (COCO). \OURS compares favorably to LLaVA-v1.5-7B and reaches similar or better performance in the benchmark compared to VLMs with much more parameters. This performance advantage can be attributed to the specialized face training dataset and the usage of our face-centric conversation. These elements enhance \OURS's proficiency in interpreting face-related language-based instructions, thereby improving its overall effectiveness in relevant tasks.

\subsection{Ablation study}
\textbf{Influence of LLM on face understanding.} Classical methods mainly rely on a vision encoder for regressing the 3DMM face parameters. In contrast, LLaVA utilizes a combination of frozen vision encoder and an LLM to perceive information from images. To study the effect of the LLM head, we train a baseline model with the vision encoder of LLaVA and an MLP to predict the human face parameters directly. The results are presented in Table \ref{ablation:effect_llm} and show that the additional LLM helps in predicting better face parameters based on the visual representations from the frozen encoder.

\textbf{Influence of face-centric conversation generation.} To prove the necessity of our face-centric conversation generation strategy, we train a model only using a simple single-turn conversation template for face images, which is a common strategy used in the VLM-based image understanding works like LISA and ChatPose. The comparison results are demonstrated in Table \ref{ablation:face_conv}. We observe that face-centric conversations help a lot in improving model's ability in performing detailed description and complex reasoning. The face-centric conversations only have a small effect on the face reconstruction task, which is expected, as the generated conversations do not contain information about the 3DMM parameters of faces. 

\textbf{Comparison between supervised face loss and unsupervised face loss.} Recent VLM-based works on segmentation and pose estimation generally rely on supervised learning with ground truth data or pseudo ground truth data. To study the influence of supervision on the model's capability, we compare the model trained with supervised 3DMM losses and self-supervised 2D face losses. For the supervised 3DMM loss, we extract the 3DMM parameters on the same CelebA-HQ trainset with the state-of-the-art face reconstruction method FOCUS \citep{li2023focus}. L1 loss is used for optimization on 3DMM ground truth. We observe in \cref{ablation:comp_sup_unsup}, that the model utilizing a supervised 3DMM loss outperforms its unsupervised counterpart, indicating that the performance ceiling is not reached yet and improvements on the self-supervised training could potentially lead to further performance gains.

\begin{table}
  \caption{GPT4-Assisted Evaluation on instruction-following capability. ``Conv", ``Details", and
``Complex" correspond to three types of questions(conversation, detailed description, complex reasoning) produced by LLaVA's data generation pipeline. GPT4 will be prompted to evaluate the answers from different models along with the ground truth answer produced by text-only GPT4. It would then give a score for each answer with an explanation. Previous method like ChatPose \citep{feng2024chatpose} uses gpt-4-0314 for evaluation which is not accessible anymore. Thus, we provide the evaluation results with a newer gpt-4-0613.}
  \label{gpt_res}
  \centering
  \begin{tabular}{l|l|cccc}
    \toprule
    % \multicolumn{2}{c}{Part}                   \\
    % \cmidrule(r){1-2}
         & Evaluation &Conv     & Detail & Complex & All         \\
    \midrule
    LLaVA-V1.5-13B & \multirow{2}{6em}{gpt-4-0314} &84.4& 81.0& 93.9 & 86.5 \\
    ChatPose \citep{feng2024chatpose}         & & 78.8 & 76.2 & 96.7 & 84.0 \\
    \midrule
        LLaVA-V1.5-13B \citep{liu2023improvedllava}&\multirow{3}{6em}{gpt-4-0613} & 80.4 & 81.4 & 90.9 & 84.2 \\
    LLaVA-V1.5-7B \citep{liu2023llava}    &       & 79.9 & 77.6 & 92.4 & 83.4 \\

    FaceGPT   & &    79.6   & 81.5      & 92.6 & 84.6\\
    \bottomrule
  \end{tabular}
  \caption{Influence of LLM towards human face understanding}
  \label{ablation:effect_llm}
  \centering
  \begin{tabular}{lll}
    \toprule
    Method     &  photo      & keypoint          \\
    \midrule
    CLIP ViT + MLP   &  0.133           & 7.6px             \\
    
    FaceGPT   &  0.106           & 3.0px                \\
    \bottomrule
  \end{tabular}
  \caption{Influence of face-centric conversation generation}
  \label{ablation:face_conv}
  \centering
  \begin{tabular}{l|c|cc|cccc}
    \toprule
    Method      & face convs & photo      & keypoint & Conv & Detail & Complex & All      \\
    \midrule
    LLaVA-1.5-7B&  \xmark    & - & - & 79.9 & 77.6 & 92.4 & 83.4 \\
    FaceGPT     &  \xmark    &  0.110  &  3.2px  & 78.4 & 80.8 & 89.0 & 82.7 \\
    FaceGPT     & \checkmark &  0.106  &  3.0px  & 79.6   & 81.5      & 92.6 & 84.6\\
    \bottomrule
  \end{tabular}
\end{table}

\begin{wraptable}{r}{5cm}
  \caption{Comparison when learning with supervised and unsupervised losses.}
  \label{ablation:comp_sup_unsup}
  \centering
  \begin{tabular}{lll}
    \toprule
    Method         &  photo. & keyp.  \\
    \midrule
    FaceGPT(sup) &0.092       &  2.1px    \\
    FaceGPT(unsup) &0.106       &  3.0px    \\
    \bottomrule
  \end{tabular}
\end{wraptable}

\section{Limitations and Future Work}
\label{limitation_part}
While \OURS demonstrates remarkable results at reasoning about 3D faces from visual and text input with unsupervised learning, many open challenges remain.
Our work represents merely an initial step into this promising area, highlighting the vast potential for future exploration and advancement in this direction.
The model does not yet match the state-of-the-art performance of task-specific 3D face reconstruction methods, and one way to address this problem is to largely scale up the training to include various face databases.
The extension of \OURS to include arbitrary numbers of faces in an image is another interesting potential research direction.
Finally, the model is specific to faces and relies on the availability of a 3D morphable model for faces. A generalization to general objects would require the self-supervised learning to also include the generative object model parameters.

\section{Conclusion}
\OURS is the first self-supervised learning framework for Large Vision-Language Models to reason about 3D human faces.
We show that VLMs can learn to predict detailed 3D human faces from not only images, but also from textual inputs, in a fully self-supervised manner via inverse rendering. 
As a generalist model, \OURS achieves strong results across various tasks, including traditional 3D face reconstruction, visual instruction following and text-based face reconstruction.
We believe our work also has general implications beyond face analytics, as it points towards a way forward to enable large multi-modal language-models to reason about our 3D world without supervision.

\newpage

\bibliography{Styles/neurips_2024}
\end{document}